\documentclass[letterpaper]{article}

\usepackage{natbib,alifeconf}  
\usepackage{amsmath}
\usepackage{url,hyperref,cleveref}
\usepackage{booktabs}

\title{More complex environments may be required to discover benefits of lifetime learning in evolving robots}

\author{
    Ege de Bruin$^{1}$,
    Kyrre Glette$^{1,2}$ \and
    Kai Olav Ellefsen$^{1}$ \\
    \mbox{}\\
    $^1$Department of Informatics, University of Oslo, Norway\\
    $^2$RITMO, University of Oslo, Norway
    \\ egedebruin@gmail.com
} 

\begin{document}

\maketitle

\begin{abstract}
    It is well known that intra-life learning, defined as an additional controller optimization loop, is beneficial for evolving robot morphologies for locomotion. In this work, we investigate this further by comparing it in two different environments: an easy flat environment and a more challenging hills environment. We show that learning is significantly more beneficial in a hilly environment than in a flat environment and that it might be needed to evaluate robots in a more challenging environment to see the benefits of learning.
    
\end{abstract}

\section{Introduction}
In evolutionary robotics, one of the main challenges is the co-evolution of morphology and control \citep{Cheney2016}. This is because we are optimizing two components of a robot at the same time, and a control setting that works for one robot morphology might not work for its offspring at all. So an evolutionary process might get stuck at a local optimum, as it is unable to find optimal control for certain robot morphologies. Adding a lifetime learning loop for the robot to optimize its control is an intuitive way to deal with this problem \citep{eiben_if_2020} and has shown good results \citep{miras_evolving-controllers_2020, luo_effects_2022} even with randomly initialized controllers \citep{zhao_robogrammar_2020, gupta_embodied_2021}. 

We investigate how adding a control learning phase to a robot's lifetime affects its performance in several types of environments, as a change in environment can produce different robots \citep{MirasEnv2020}, even though that is not as trivial as might be expected\citep{MirasEnv2019}. The robot's morphology is evolved using an evolutionary algorithm, while its control is optimized using Bayesian Optimization \citep{lan_learning_2021} with three different learning budgets: one where there is no control optimization done and the parameters are fully random, and learning budgets of 30 and 50 iterations. We compare two types of environments, an "easy" flat environment and a more complex hilly environment, and want to see how environment complexity relates to control learning. The main contribution of this work is that lifetime learning is more beneficial in a more challenging environment than in a completely flat environment.

    

\section{Methods}

\subsection{Phenotype}
For this work, we make use of Revolve2 \footnote{See https://github.com/ci-group/revolve2}{} as a modular robot framework. The morphologies are built using three different building blocks, namely a core module, a brick module, and a hinge module. Every robot has exactly one core module.

The control of the robot is a decentralized approach, where every hinge has its sine wave to control its angle. Every hinge also has a touch sensor that is used as input for controlling its and its neighbours' angles. The following equations are used for control:
\begin{equation}
\begin{split}
\Theta & = A * sin(\phi + P) + O \\ \phi & = \phi + \Delta_\phi * F \\ S_1 & = s * W * sin(\phi + O) \\ S_2 & = s_{N} * W_{N} * sin(\phi + O)
\end{split}
\end{equation}

In these equations, $\Theta$ is the output for the controller, $A$ is the amplitude, $P$ is the phase offset and $F$ is the frequency. The input of a hinge's touch sensor is $s$, which is a binary value, and $W$ is the weight of the touch sensor. The input of neighbouring hinge's touch sensors, where a neighbour is defined as a hinge within a manhattan distance of 2, is $s_N$, which is a binary value as well that is 1 if at least one neighbour's touch sensor is activated, and $W_{N}$ is the weight of the neighbour touch sensor. Finally, $O$ is the phase offset when a touch sensor is activated. $A, P, W, W_{N}$ and $O$ are the five learnable parameters, and $F$ is set to 4. Even though the control is decentralized, hinges might share parameters. We will experiment with different numbers of sets of control parameters, that can be used by the hinges.

\subsection{Genotype}
The robot morphology is directly encoded in the genotype. Every robot has a core module and the first robots are initialized by adding several random modules in random possible slots iteratively. If a hinge module is added, it chooses one of the possible controller parameters randomly. Crossover between two parents is done by swapping the modules on one random location of the core module of the two robots, resulting in two offspring robots. Every offspring then goes through mutation, which can be one of three possible mutations. Firstly, a random number of modules can be added to the robot. Secondly, a random number of modules can be removed from the robot, and if a non-leaf module is removed, all modules starting from it are removed as well. Finally, the hinges can switch controllers. The control of the robot is not influenced during evolution and therefore has no genotype.

\subsection{Evolution and Learning}
All experiments will evolve the robot morphology, where every morphology will go through a learning phase to find the best control parameters. The morphology evolution loop is a standard evolutionary algorithm, with tournament selection as survivor and parent selection. For the survivor selection, both the original population and offspring population are evaluated in the tournaments. 

The learning algorithm is Bayesian Optimization, where controller parameters are learned from scratch, meaning there is no controller inheritance. We use the Matern 5/2 kernel with a length scale of 0.2, and the Upper Confidence Bound as an acquisition function with an exploration variable of 3, which has been used and worked well before \citep{lan_learning_2021, vanDiggelen2021}. Before starting the Bayesian Optimization, 10\% of the learning budget will be used to generate random samples, these are sampled using Latin Hypercube Sampling. We experiment with different numbers of learning iterations, namely 1, 30, and 50. A learning budget of 1 means that there is no Bayesian Optimization used and a robot morphology's performance is based on only a single run. In this case, the control parameters are fully random at each evaluation.

\begin{figure}
    \centering
    \includegraphics[width=0.9\linewidth]{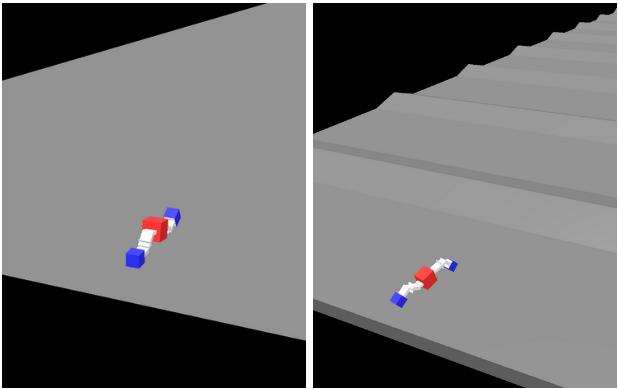}
    \caption{
        An example robot in the two environments. Left is the flat environment and right is the hills environment.
    }
    \label{figure-envs}
\end{figure}

\subsection{Simulation}
As Revolve2 is used as the framework for robot evolution, the robots are simulated using MuJoCo. The robots are evaluated on their movement in a certain direction, whereas fitness is the distance moved in that direction. To study the effect of environmental complexity, we experiment with two different environments. The first environment is flat and has no obstacles. The second environment is a hilly environment, where every two meters there is a hill of 0.35 meters for the robot to climb over, see Figure \ref{figure-envs}.

The number of generations is 9995 for the 1 learning budget experiment and 495 for the others, this way we can compare both generations and function evaluations more fairly. The initial number of modules is random between 5 and 10, and the simulation time per evaluation is 30 seconds. There are three learning budgets (1, 30, and 50), three different numbers of controllers (1, 4, and 8), and two environments (flat and hills), giving 18 experiments in total. All experiments are run 5 times.


\begin{figure}
    \centering
    \includegraphics[width=1\linewidth]{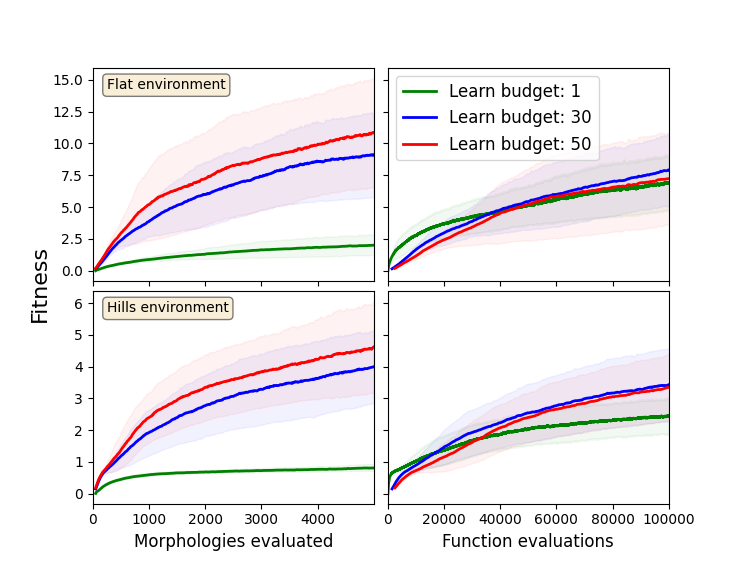}
    \caption{
        Plots comparing the fitness of different learning budgets, left plots on the \textbf{flat} environment and right plots on the \textbf{hills} environment. The line shows the mean fitness of the population after the number of morphologies evaluated/function evaluations, combining all runs for that learning budget. Shaded areas are the standard deviation.
    }
    \label{figure-line-both}
\end{figure}

\section{Results}

To see how adding a controller learning phase affects the performance, we compare the fitness over the number of morphologies evaluated, and over the number of function evaluations. Comparing on morphologies makes sense if the cost of creating a robot is higher than evaluating a robot and visa versa. The top two plots of Figure \ref{figure-line-both} show the results on the flat environment. When comparing the fitness over morphologies, it is clear that the runs with a single learning evaluation perform relatively poorly. This makes sense, as experiments with a higher learning budget have more function evaluations per morphology to find better control parameters. This is why we also plot the fitness over function evaluations. Initially, the runs with a learning budget of 1 do perform better. However, after 100.000 function evaluations, the differences are minimal. Note that the 1-learn experiments evaluate more morphologies than 5000, and the other learn experiments have more than 10.000 function evaluations, but for comparison, the graphs are cut off at 5.000 morphologies evaluated and 10.000 function evaluations.

In the bottom two plots of Figure \ref{figure-line-both} the same plots are shown, but then in the hills environment. The plot of fitness over morphologies evaluated shows similar results as the flat environment, however, when comparing on function evaluations there is a difference. Towards the end, there is a clear difference between the runs with a learning budget of 1 and the other learning budgets. This shows that a more complex environment might need controller optimization over a robot's lifetime to perform well after a certain number of generations. 

\begin{figure}
    \centering
    \includegraphics[width=\linewidth]{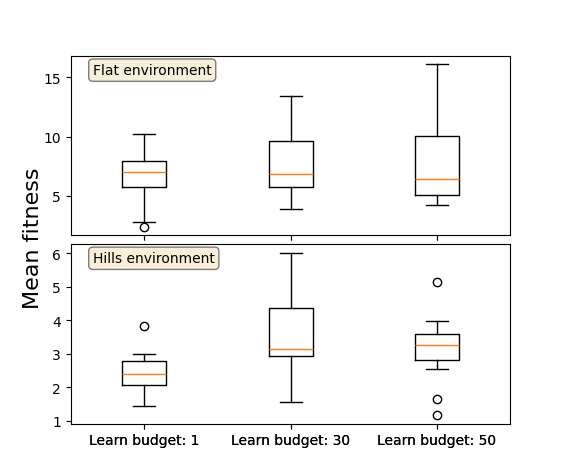}
    \caption{
        Box plots showing the mean fitness of the generation after 100.000 function evaluations.
    }
    \label{figure-box-fe}
\end{figure}

The boxplots in Figure \ref{figure-box-fe} confirm the differences. The figure shows the mean fitness of the generation after 100.000 function evaluations. A Wilcoxon test shows a significant difference between a learning budget of 1 and 50 on the hills environment after 100.000 function evaluations (p $<$ 0.05). There is a more significant difference between a learning budget of 1 and 30 on the hills environment after 100.000 function evaluations (p $<$ 0.01).


\section{Conclusion}
We compared different learning budgets when evolving morphology and control on two types of terrains and showed that learning control is more beneficial in a more challenging environment. That learning is more beneficial is visible when comparing on how many morphologies are evaluated, but when comparing on function evaluations the difference is more explicit in a more challenging environment than in an easier environment. In a flat environment, morphology evolution seems to solve the problem of locomotion without the need for learning, and to study the effect of intra-life learning a more challenging environment is needed.

For future work, we would like to compare the results to an evolutionary search with no learning, as currently there is no evolutionary search on controller parameters and this might improve performance in the no-learning case significantly. This can also be extended by adding it to the learning experiments, adding Lamarckian inheritance to the robots' control. There is also some work to be done on optimizing the robots' control, for example optimizing the sine wave equations, removing unnecessary parameters, and optimizing the number of controllers a robot can choose from to find a good balance between complexity and reusability.

\footnotesize

\begin{thebibliography}{}

\bibitem[Cheney et~al., 2016]{Cheney2016}
Cheney, N., Bongard, J., Sunspiral, V., and Lipson, H. (2016).
\newblock On the {Difficulty} of {Co}-{Optimizing} {Morphology} and {Control} in {Evolved} {Virtual} {Creatures}.
\newblock {\em Proceedings of the Artificial Life Conference 2016 (ALIFE XV)}, pages 226--234.

\bibitem[Eiben and Hart, 2020]{eiben_if_2020}
Eiben, A.~E. and Hart, E. (2020).
\newblock If it evolves it needs to learn.
\newblock In {\em Proceedings of the 2020 {Genetic} and {Evolutionary} {Computation} {Conference} {Companion}}, pages 1383--1384, Cancún Mexico. ACM.

\bibitem[Gupta et~al., 2021]{gupta_embodied_2021}
Gupta, A., Savarese, S., Ganguli, S., and Fei-Fei, L. (2021).
\newblock Embodied intelligence via learning and evolution.
\newblock {\em Nature Communications}, 12(1):5721.

\bibitem[Lan et~al., 2021]{lan_learning_2021}
Lan, G., De~Carlo, M., Van~Diggelen, F., Tomczak, J.~M., Roijers, D.~M., and Eiben, A. (2021).
\newblock Learning directed locomotion in modular robots with evolvable morphologies.
\newblock {\em Applied Soft Computing}, 111:107688.

\bibitem[Luo et~al., 2022]{luo_effects_2022}
Luo, J., Stuurman, A.~C., Tomczak, J.~M., Ellers, J., and Eiben, A.~E. (2022).
\newblock The {Effects} of {Learning} in {Morphologically} {Evolving} {Robot} {Systems}.
\newblock {\em Frontiers in Robotics and AI}, 9:797393.

\bibitem[Miras et~al., 2020a]{miras_evolving-controllers_2020}
Miras, K., De~Carlo, M., Akhatou, S., and Eiben, A.~E. (2020a).
\newblock Evolving-{Controllers} {Versus} {Learning}-{Controllers} for {Morphologically} {Evolvable} {Robots}.
\newblock In {\em Applications of {Evolutionary} {Computation}}, volume 12104, pages 86--99. Springer International Publishing, Cham.

\bibitem[Miras and Eiben, 2019]{MirasEnv2019}
Miras, K. and Eiben, A.~E. (2019).
\newblock Effects of environmental conditions on evolved robot morphologies and behavior.
\newblock In {\em Proceedings of the Genetic and Evolutionary Computation Conference}, GECCO ’19. ACM.

\bibitem[Miras et~al., 2020b]{MirasEnv2020}
Miras, K., Ferrante, E., and Eiben, A.~E. (2020b).
\newblock Environmental influences on evolvable robots.
\newblock {\em PLOS ONE}, 15(5):e0233848.

\bibitem[van Diggelen et~al., 2021]{vanDiggelen2021}
van Diggelen, F., Ferrante, E., and Eiben, A.~E. (2021).
\newblock Comparing lifetime learning methods for morphologically evolving robots.
\newblock In {\em Proceedings of the Genetic and Evolutionary Computation Conference Companion}, GECCO ’21. ACM.

\bibitem[Zhao et~al., 2020]{zhao_robogrammar_2020}
Zhao, A., Xu, J., Konaković-Luković, M., Hughes, J., Spielberg, A., Rus, D., and Matusik, W. (2020).
\newblock {RoboGrammar}: graph grammar for terrain-optimized robot design.
\newblock {\em ACM Transactions on Graphics}, 39(6):1--16.

\end{thebibliography}

\end{document}